\documentclass[sigconf]{acmart}

\AtBeginDocument{
  }

\usepackage{subcaption}
\usepackage[font=small,skip=3pt]{caption}
\usepackage{multirow}
\usepackage{tikz}
\usepackage{stmaryrd}
\usepackage{balance}

\copyrightyear{2026}
\acmYear{2026}
\setcopyright{cc}
\setcctype{by}
\acmConference[WSDM '26]{Proceedings of the Nineteenth ACM International Conference on Web Search and Data Mining}{February 22--26, 2026}{Boise, ID, USA}
\acmBooktitle{Proceedings of the Nineteenth ACM International Conference on Web Search and Data Mining (WSDM '26), February 22--26, 2026, Boise, ID, USA}
\acmPrice{}
\acmDOI{10.1145/3773966.3779391}
\acmISBN{979-8-4007-2292-9/2026/02}

\begin{document}

\title{Abacus: Self-Supervised Event Counting-Aligned Distributional Pretraining for Sequential User Modeling}

\author{Sullivan Castro}
\authornote{The work was done during the master internship at Criteo AI Lab.}
\affiliation{%
  \institution{Criteo AI Lab}
  \city{Paris}
  \country{France}
}
\affiliation{%
  \institution{École Nationale des Ponts et Chaussées}
  \city{Paris}
  \country{France}
}
\email{sullivan.castro75@gmail.com}

\author{Artem Betlei}
\affiliation{%
  \institution{Criteo AI Lab}
  \city{Grenoble}
  \country{France}}
\email{a.betlei@criteo.com}

\author{Thomas Di Martino}
\affiliation{%
  \institution{Criteo AI Lab}
  \city{Paris}
  \country{France}}
\email{t.dimartino@criteo.com}

\author{Nadir El Manouzi}
\affiliation{%
  \institution{Criteo AI Lab}
  \city{Paris}
  \country{France}}
\email{n.elmanouzi@criteo.com}

\renewcommand{\shortauthors}{Castro et al.}

\begin{abstract}
Modeling user purchase behavior is a critical challenge in display advertising systems, necessary for real-time bidding. The difficulty arises from the sparsity of positive user events and the stochasticity of user actions, leading to severe class imbalance and irregular event timing.
Predictive systems usually rely on hand-crafted ``counter'' features, overlooking the fine-grained temporal evolution of user intent. 
Meanwhile, current sequential models extract 
direct sequential signal,
missing useful event-counting statistics. 
We enhance deep sequential models with self-supervised pretraining strategies for display advertising. 
Especially, we introduce Abacus, a novel approach of predicting the empirical frequency distribution of user events. 
We further propose a hybrid objective unifying Abacus with sequential learning objectives, combining stability of aggregated statistics with the sequence modeling sensitivity. 
Experiments on two real-world datasets show that Abacus pretraining outperforms existing methods accelerating downstream task convergence, while hybrid approach yields up to +6.1\% AUC compared to the baselines.
\end{abstract}

\maketitle

\section{Introduction}

Modeling online user behavior is central to display advertising (DA), where Demand-Side Platforms (DSPs) must decide within milliseconds whether and how much to bid for an impression. The setting is challenging: purchase events are extremely rare relative to clicks and impressions, and event timing is highly irregular, with long and variable gaps between user actions.

In practice, production systems rely heavily on handcrafted, count-based features - aggregated statistics over fixed time windows~\citep{pinterest2018aperture, zhou2018atrank, snap2022robusta, singh2023adload}. These counters are robust and capture coarse patterns, but they under-represent the fine-grained temporal evolution of intent, motivating direct sequence modeling. Deep sequential models have proven their effectiveness in recommendation~\citep{kang2018self, sun2019bert4rec, zhou2020s3}, yet in DA they are often inferior to systems built on counting features, providing stable aggregations
~\citep{du2023frequency, abbattista2024enhancing, han2024enhancing, fang2024general}.

A common remedy across domains is self-supervised learning (SSL) pretraining before finetuning~\citep{devlin2019bert, brown2020language, chen2020simple}, which brings (i) label efficiency when positives are scarce, and (ii) transferable representations. However, standard SSL tasks (e.g. next item prediction, masked modeling, contrastive learning) do not explicitly align with the counting signals that drive performance.

We introduce \textsc{Abacus}, a counting-aligned pretext task that trains a sequential encoder to predict the empirical event-type distribution within a user timeline. Abacus supplies stable supervision that reflects the information captured by production counters (without manual feature engineering) and complements token-level pretexts. We further propose a hybrid objective that mixes Abacus with masked modeling \citep{devlin2018bert} and Barlow Twins \citep{zbontar2021barlow}, combining the robustness of aggregation with the sensitivity of temporal encoders.

On a large private dataset, Abacus-based pretraining improves AUC by up to +6.1\% over a scratch-trained GRU encoder, with smaller but consistent gains on the public Taobao benchmark. Besides, Abacus yields a warm-start effect and faster convergence. Ablations show that mixing Abacus with complementary pretext tasks is beneficial and stable.
\vspace{-1em} 
\paragraph{Contributions.}
We (i) show that self-supervised pretraining on event sequences improves user behavior modeling for DA, improving performance and accelerating convergence versus training from scratch, (ii) introduce \textsc{Abacus}, a counting-aligned objective that predicts a sequence’s empirical event distribution (including standard, random, and masking variants) and (iii) propose a hybrid, multi-task approach that mixes Abacus with masking and contrastive pretext tasks, improving accuracy and stability across used encoders.

\section{Related Work}





Early approaches to user modeling for DA relied on LSTM~\citep{zolna2017user, lang2017understanding} or hybrid~\citep{gligorijevic2019time} encoders of user activity sequences that captured temporal dependencies in click or browsing logs to make direct predictions. 
BERT4Rec~\citep{sun2019bert4rec}, architecture for sequential recommendation, demonstrated combination of bidirectional attention with pretraining via masked modeling to be efficient to model user behavior. Later on,~\citep{zhao2022resetbert4rec} extended the method by combining masked modeling with sequence rearrangement, explicitly integrating relative time embeddings.
In Pretrained User Models (PTUM)~\citep{wu2020ptum}, two pretext tasks for user model pretraining from large-scale unlabeled data are proposed: the model learns relatedness between historical behaviors with masked behavior prediction, while the relatedness between past and future behaviors is captured by next K behavior prediction. UserBERT~\citep{wu2022userbert} complemented PTUM by suggesting pretraining by behavior sequence matching, which aims to capture inherent user interests that are consistent in different periods.
\citep{liao2020effectiveness} explored benefits of SSL pretraining for event-based user behavior sequences. Their pretraining approach of contrasting the GRU-encoded representations of the possible future events from those of the less likely events with a contrastive loss has demonstrated good performance on the downstream conversion prediction task.
\citep{pancha2022pinnerformer} developed PinnerFormer, a Transformer-based model tailored for long user engagement sequences on Pinterest. Unlike methods focusing on next-item prediction, PinnerFormer employs a dense all-action loss that jointly predicts multiple future engagement types (e.g., repins, close-ups, link clicks) across long horizons, demonstrating strong performance in capturing evolving user interests for large-scale recommendation.
\citep{fu2023robust} proposed taking the prediction of user behavior distribution over a defined period in the future as a pretext task, instead of the usual next K behaviors prediction. They provided the Multi-scale Stochastic Distribution Prediction (MSDP) algorithm, improving the generalization ability of user representation and satisfying various downstream tasks. While the idea of predicting behavior distribution is similar, MSDP aims to extract sequential patterns to predict \textit{future} actions, differing from our approach of learning \textit{present} event statistics.
\citep{liu2025enhancing} adapted Barlow Twins - contrastive SSL method based on minimizing cross-correlation between pairs of differently corrupted inputs - to user sequence modeling, proven to learn informative representations without relying on labeled data or negative sampling.


\section{Method}
\begin{figure*}[th!]
  \centering

  \includegraphics[width=0.75\linewidth]{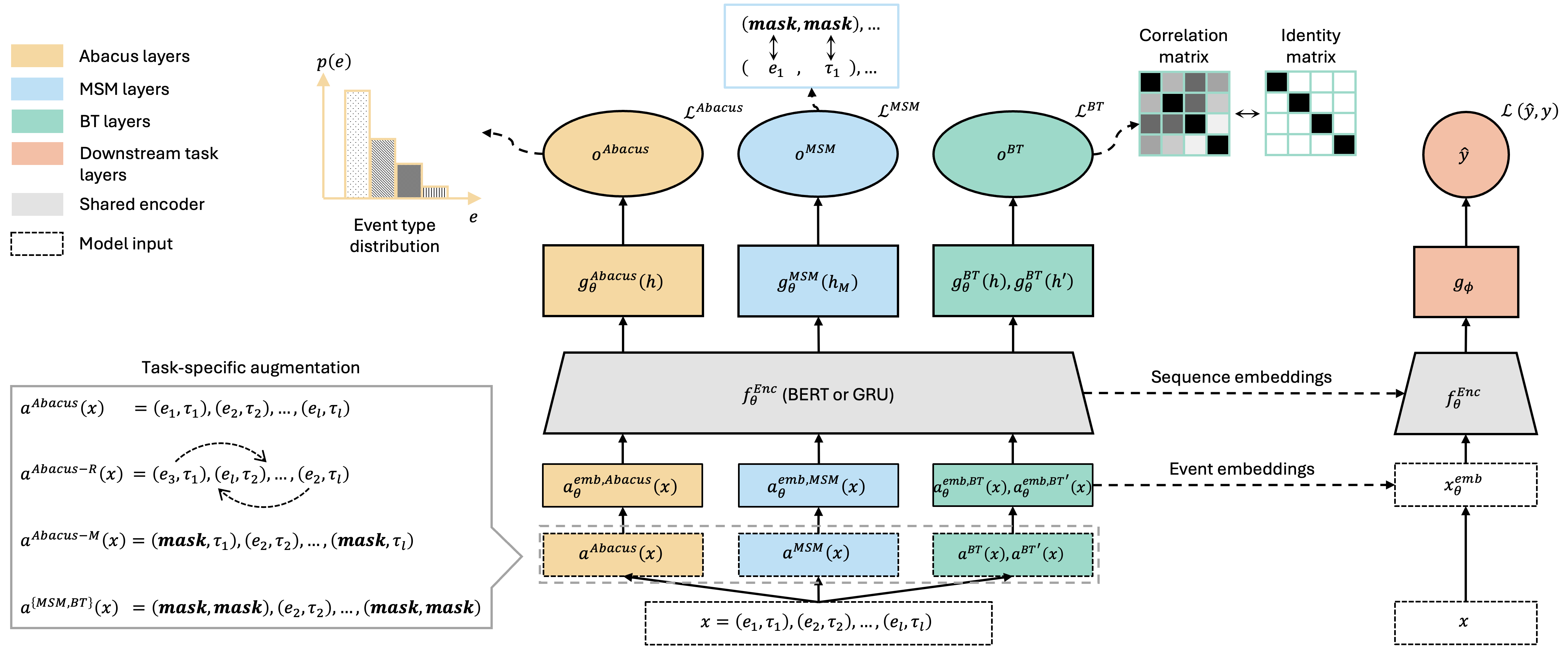}

  \caption{\small Illustration of the proposed approach. In pretraining, Abacus, MSM and BT pretext tasks are jointly learned via MTL. Both event and sequence embeddings are then finetuned on downstream task.}
  \label{fig:method_fig}
\end{figure*}


\subsection{Notation}
Let the user behavior sequence be $x = \{(e_i,\tau_i)\}_{i=1}^{\ell}$ with event type $e_i\in\{1,\dots,K\}$ and continuous normalized event timestamp $\tau_i\in [0, 1]$, $\ell$ is the sequence length. Pretrained and finetuned model parameters are defined as $\theta$ and $\phi$ respectively.
Also, let event embedding be $e_{\theta} \in \mathbb{R}^{K\times d}$ where $d$ is event embedding dimension, sequential encoder be $f^{Enc}_\theta$ and sequence embedding be $h$. 
Then, pretext task-specific entities are defined: sequence augmentation, encoder projection head, output and loss are $a^t (x)$,
${g}^t_{\theta}$, ${o^t}$ and $\mathcal{L}^t$ respectively, where $t$ denotes pretext task. 
For downstream task model, prediction head is denoted as $g_\phi$.
\subsection{Pretraining}

On the pretraining stage, we first apply a task-specific augmentation $a^t$ to the raw sequence $x$, yielding $a^t(x)$. 
Input representation is a concatenation of the learnable event embedding $(e_{\theta})_i \in \mathbb{R}^d$ with $\tau_i$, i.e. $a^{emb,t}_\theta(x) = \big\{\, {(e_{\theta})}_i \;\Vert\; \tau_{i} \,\big\}_{(e,\tau)\in a^t(x), i \in \{1,...,\ell\}}$.

\emph{Shared} encoder $f^{Enc}_\theta$ maps $a^{emb,t}_\theta(x)$
to the $h = f^{Enc}_\theta \big(a^{emb,t}_{\theta}(x) \big)$ embedding summarizing a sequence. For the encoder architecture, two models are considered: RNN (GRU) and Transformer (BERT with [CLS] token at the end as a summary, as done in related works \citep{wang2023bert4ctrefficientframeworkcombine}).
Then, the task-specific head $g^t_{\theta}$ maps $h$ to the prediction ${o^t} = g^{t}_{\theta} (h)$, forming the loss $\mathcal{L}^t$.

For the hybrid model training, multi-task learning (MTL) is applied: on a mini-batch, task-specific losses are computed on the \emph{same} examples, so single backward pass is performed on the weighted sum: $\mathcal{L}^{\text{MTL}}=\sum_{t} w^t\,\mathcal{L}^t$ where $w^t\!\ge\!0$ and $\sum_t w^t=1$.

Chosen pretext tasks with corresponding augmentations, projection heads, and losses are described below.

\subsubsection{Pretext tasks}
\label{sec:pretext_tasks}
Three pretext tasks are used in our approach.
\paragraph{(1) Distributional (Abacus)}
Abacus trains the encoder to summarize a sequence by its \emph{global} event composition.
Given $x$, let $p^{Abacus}=\frac{1}{l}\sum_{i=1}^{\ell}\mathrm{one\_hot}(e_{i})\in \Delta^{K-1}$ where  $\Delta^{K-1}=\{p \in [0, 1]^K: \sum_{k=1}^K p_k = 1\}$ be the empirical histogram of event types.
A MLP head $g^{Abacus}_{\theta} (h)$ outputs logits histogram prediction $o^{Abacus} = g^{Abacus}_{\theta}(h)$ and $\hat{p}^{Abacus} = \text{softmax}(o^{Abacus})$.
Cross-entropy loss $\mathcal{L}^{\text{Abacus}} = -\sum_{k=1}^{K} {p}^{Abacus}_k\,\log {\hat p}^{Abacus}_{k}$ is minimized.
The choice of histogram prediction rather than absolute count is due to  sequence length invariance and compatibility across users.


Three sequence augmentation options are considered:

\begin{itemize}
  \item $a^{Abacus} (x)$ (no augmentation): learns event frequencies.
  \item $a^{Abacus-R} (x)$ (random permutation): enforces \emph{permutation invariance}, acting as additional regularizer.
  \item $a^{Abacus-M} (x)$ (segment event masking): predicts histogram from the corrupted sequence where events are masked for some segments. By this, the encoder is forced to learn how local fragments inform aggregate behavior, i.e. how observed signals correlate with unobserved ones.
\end{itemize}
Overall, Abacus injects a counting-aware prior that is order/length-agnostic and robust to imbalance.

Besides proposed Abacus task, we introduce to the model two additional, existed tasks, focused on temporal pattern learning:

\paragraph{(2) Masked Sequence Modeling (MSM)}
In MSM, both events and timestamps are masked in contiguous sequence segments $M$, forming $a^{MSM} (x)$. Then, projection head $g^{MSM}_{\theta}$ inputs $h_M$ (masked elements of $h$ only) and outputs
$o^{MSM} = \big( o^{MSM}_{(1:K)}, o^{MSM}_{(K+1)} \big)$
- so $o^{MSM}_{(1:K)}$ are event reconstruction logits, and the last component is a timestamp reconstruction $\hat \tau = o^{MSM}_{(K+1)}$.
Then we can derive $\hat{p}^{MSM} = \text{softmax}(o^{MSM}_{(1:K)})$. For the event and timestamp reconstruction, cross-entropy and (MSE) losses are used respectively:
\[
\mathcal{L}^{\text{MSM}}
=-\frac{1}{|M|} \sum_{m \in M} \,\log {\hat p}^{MSM}_{e_m}
+\lambda^{MSM} \frac{1}{|M|}\sum_{m \in M}\big(\tau_m-\hat{\tau}_m\big)^2.
\]




$\lambda^{MSM} = 1$ is used for the experiments. MSM trains the encoder to recover locally coherent event-time patterns, sharpening short-range temporal context and acting as a denoising regularizer, so the learned representations are more robust to missing or corrupted spans providing a stronger warm-start for finetuning.

\paragraph{(3) Barlow Twins (BT)}
Two identical inputs with independently masked segments, $a^{BT} (x)$ and ${a^{BT}}'(x)$, are forwarded to $f^{Enc}_\theta$ producing $h$ and ${h}'$, then, projection $g^{BT}_\theta$ outputs ${o^{BT}}(h, {h}') = (z, {z}')$.


Cross-correlation matrix of the normalized outputs $(\tilde{z}^i, \tilde{z'}^j)$ where $\tilde z = \frac{z-\mu}{\sigma}$ (with mean $\mu$, standard deviation $\sigma$) is $C_{ij} = \frac{\langle \tilde z^i, \,\tilde z'^{j} \rangle}
{\|\tilde z^{i}\| \|\tilde z'^{j}\|}$.
The Barlow Twins loss is $\mathcal{L}_{\text{BT}} = \sum_{i} (1 - C_{ii})^2 + \lambda_{BT} \sum_{i}\sum_{j \neq i} C_{ij}^2,$
which enforces invariance ($C_{ii}\!\approx\!1$) while reducing redundancy between dimensions ($C_{ij}\!\approx\!0$). $\lambda_{BT} = 1$ has been used.


\subsection{Finetuning}
In the finetuning stage, both the event embeddings $e_\theta$ and encoder $f^{Enc}_\theta$ are finetuned on downstream task, while projection MLP head $g_\phi$ is learned from scratch, returning $\hat y$, as seen in Figure \ref{fig:method_fig}.

\section{Experiments}

\subsection{Datasets}
\label{sec:datasets}
We evaluate our method on two real-world, large-scale
datasets of (event type, timestamp) pair sequences: 
\begin{itemize}
    \item \textbf{Taobao}\footnote{\url{https://tianchi.aliyun.com/dataset/649}}: user behavior log from the big chinese e-commerce platform, collected during 8 days and containing ``click'', ``add-to-cart'', ``add-to-fav'' and ``purchase''  events.
    \item \textbf{Private}: user behavior log from online advertising platform, collected on one partner during one month and containing 9 user event types.
\end{itemize}

\begin{table}[h!]
\centering
\caption{Dataset statistics (higher is better for both PPL and GS).}
\label{tab:dist-diagnostics}
\small
\resizebox{0.8\columnwidth}{!}{%
\begin{tabular}{l|cccccc}
\toprule
Dataset & Size & Seq. Length & Label Mean & PPL $\uparrow$ & GS $\uparrow$ \\
\midrule

Taobao  &   1M & 100 & 0.0131 & 1.57 & 0.20 \\
Private & 1.7M &  50 & 0.1014 & 4.43 & 0.71 \\

\bottomrule
\end{tabular}
}
\end{table}

To qualitatively compare the datasets, we report general numbers (size, sequence length used and label mean) along with the event distribution statistics: i) \emph{Perplexity} $\text{PPL}=2^{-\sum_i p_i \log_2 p_i}$, representing the effective number of equally likely events, and ii) \emph{Gini-Simpson diversity} $\text{GS}=1-\sum_i p_i^2$ which is the probability that two independent draws differ. Statistics are provided in Table \ref{tab:dist-diagnostics}. Private dataset exhibits greater event type balance and label density than Taobao.

As we consider purchase prediction to be the downstream task, for both datasets each sequence is split by time into i) a user history $x$, used for training, and ii) labeling time window. We assign label $y = 1$ if a purchase event occurs at least once in this window, otherwise $y = 0$. 
70/20/10 and 60/20/20 time-based train/val/test split are used for Private and Taobao respectively.

\subsection{Experimental Setup}

\paragraph{Baselines}
We use two types of baselines: i) \textit{No Pretraining (No-PT)} so finetuning only, to check the performance gain due to pretraining stage, and ii) baseline pretrainers including \textit{Next Event Prediction (NEP)}, \textit{Next K Events Histogram Prediction (NKEHP)} \citep{fu2023robust}, \textit{Masked Sequence Modeling (MSM)}, and \textit{Barlow Twins (BT)} \citep{liu2025enhancing} to compare to both Abacus and hybrid (MTL) pretraining approaches.

All experiments were done in PyTorch on 2 NVIDIA V100 (16 GB) GPUs. Dimensions of $e_\theta$ and $h$ embeddings are 3 and 8 respectively. Batch size is $2^{14}$. Learning rates were chosen by a small grid on validation AUC on 300 epochs with early stopping: 0.01 for pretraining and 0.01 and 0.001 for finetuning on Private and Taobao respectively. Models share the same tuning budget. AdamW optimizer was used. The code is publicly released\footnote{\url{https://anonymous.4open.science/r/Abacus/}}.



\subsection{Results}
\label{sec:results}
For both datasets, we report \emph{mean $\pm$ std} AUC over 3 random seeds computed on the test split, along with the relative improvement over No-PT. Best results are highlighted in bold.

\paragraph{Taobao dataset (Table~\ref{tab:taobao-results})}
While GRU is stronger without pretraining, BERT benefits substantially more from self-supervised objectives.
Naïve event-level tasks such as NEP or MSM degrade performance across both architectures; in contrast, BT offers small but consistent improvements. Abacus variants yield the most robust single-task gains. Finally, the hybrid MTL approach combining Abacus, MSM, and BT, achieves the best overall performance.


\paragraph{Private dataset (Table~\ref{tab:criteo-results})}
Among single pretraining objectives, results are mixed: NEP significantly boosts GRU but fails for BERT, while MSM shows the opposite pattern, hurting GRU slightly but clearly improving BERT. Barlow Twins again yields modest but consistent gains for GRU while being neutral for BERT. In contrast, the Abacus family demonstrates robust improvements for both architectures, with Abacus-R being strongest. The hybrid MTL model achieves the best performance overall.
\begin{table}[h!]
\centering
\caption{\small Results on Taobao dataset.} 
\label{tab:taobao-results}
\setlength{\tabcolsep}{5pt}
\footnotesize
\begin{tabular}{lcccc}
\toprule
\multirow{2}{*}{\textbf{Model}} & \multicolumn{2}{c}{\textbf{GRU}} & \multicolumn{2}{c}{\textbf{BERT}} \\
\cmidrule(lr){2-3}\cmidrule(lr){4-5}
& \textbf{AUC}$\uparrow$ & \textbf{$\Delta$ (\%)}$\uparrow$ & \textbf{AUC}$\uparrow$ & \textbf{$\Delta$ (\%)}$\uparrow$ \\
\midrule
No-PT        & 0.6423 $\pm$ 0.0031 & +0.00 & 0.6077 $\pm$ 0.0378 & +0.00 \\
\hline
NEP          & 0.5927 $\pm$ 0.0382 & $-$7.72 & 0.6170 $\pm$ 0.0295 & +1.53 \\
NKEHP        & 0.6342 $\pm$ 0.0076 & $-$1.26 & 0.6133 $\pm$ 0.0318 & +0.92 \\
MSM          & 0.6409 $\pm$ 0.0008 & $-$0.21 & 0.6073 $\pm$ 0.0448 & $-$0.07 \\
BT           & 0.6423 $\pm$ 0.0006 & +0.01  & 0.6024 $\pm$ 0.0474 & $-$0.87 \\
\hline
Abacus       & 0.6422 $\pm$ 0.0009 & $-$0.01 & 0.6347 $\pm$ 0.0058 & +4.43 \\
Abacus-R     & 0.6437 $\pm$ 0.0017 & +0.22 & 0.6326 $\pm$ 0.0074 & +4.09 \\
Abacus-M     & 0.6421 $\pm$ 0.0015 & $-$0.03 & 0.6310 $\pm$ 0.0032 & +3.83 \\
\hline
Hybrid (MTL)$^{\dagger}$ & \textbf{0.6439} $\pm$ \textbf{0.0010} & \textbf{+0.25} & \textbf{0.6411} $\pm$ \textbf{0.0017} & \textbf{+5.49} \\
\bottomrule
\end{tabular}
{\\\scriptsize
$\dagger$ Weights: GRU 0.6/0.4 (Abacus-R/Barlow), 
BERT 0.5/0.25/0.25 (Abacus-R/MSM/BT).
}
\end{table}
\vspace{-1em} 

\begin{table}[h!]
\centering
\caption{\small Results on private dataset.}
\label{tab:criteo-results}
\setlength{\tabcolsep}{5pt}
\footnotesize
\begin{tabular}{lcccc}
\toprule
\multirow{2}{*}{\textbf{Model}} & \multicolumn{2}{c}{\textbf{GRU}} & \multicolumn{2}{c}{\textbf{BERT}} \\
\cmidrule(lr){2-3}\cmidrule(lr){4-5}
& \textbf{AUC}$\uparrow$ & \textbf{$\Delta$ (\%)}$\uparrow$ & \textbf{AUC}$\uparrow$ & \textbf{$\Delta$ (\%)}$\uparrow$ \\
\midrule
No-PT        & 0.7094 $\pm$ 0.0149 & +0.00  & 0.7175 $\pm$ 0.0224 & +0.00  \\
\hline
NEP     & 0.7215 $\pm$ 0.0089 & +1.71  & 0.7173 $\pm$ 0.0206 & $-$0.03 \\
NKEHP   & 0.7003 $\pm$ 0.0096 & $-$1.28  & 0.7198 $\pm$ 0.0186 & +0.33 \\
MSM         & 0.7076 $\pm$ 0.0088 & $-$0.25 & 0.7304 $\pm$ 0.0086 & +1.80 \\
BT         & 0.7258 $\pm$ 0.0108 & +1.66  & 0.7164 $\pm$ 0.0141 & $-$0.13 \\
\hline
Abacus        & 0.7342 $\pm$ 0.0196 & +3.49  & 0.7324 $\pm$ 0.0149 & +2.08  \\
Abacus-R    & 0.7358 $\pm$ 0.0171 & +3.72  & 0.7378 $\pm$ 0.0058 & +2.82  \\
Abacus-M   & 0.7101 $\pm$ 0.0096 & +0.09  & 0.7214 $\pm$ 0.0158 & +0.55  \\
\hline
Hybrid (MTL)$^{\dagger}$ & \textbf{0.7525} $\pm$ \textbf{0.0036} & \textbf{+6.07} & \textbf{0.7401} $\pm$ \textbf{0.0185} & \textbf{+3.15} \\
\bottomrule
\end{tabular}
{\\\scriptsize
$\dagger$ Weights: GRU 0.75/0.25 (Abacus-R/BT), BERT 0.7/0.3 (Abacus-R/BT).
}
\end{table}

\begin{figure}[h!]
  \centering
  \includegraphics[width=0.9\columnwidth]{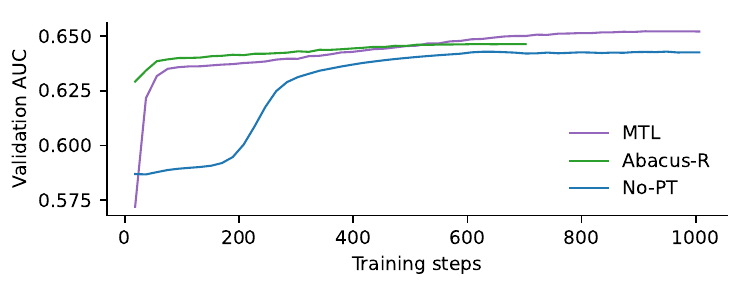}
    \captionof{figure}{\small BERT on Taobao}
    \label{fig:bert_val_auc_curves}
\end{figure}

Figure~\ref{fig:bert_val_auc_curves} shows that Abacus-pretrained models start from higher validation AUC and converge faster. Pretraining provides consistently higher validation performance, warm-start effect, and faster convergence, thus confirming existing observations~\citep{liao2020effectiveness}. The gap is largest early in training and narrows over time without collapsing, suggesting accelerated optimization and a shifted convergence point. MTL model has learnt more complex features, converging slower though reaching higher final AUC. 

\subsection{Discussion}

\textit{1. Distributional pretraining is effective.}
Counting-based objectives consistently outperform reconstruction and contrastive pretraining, indicating that aligning with distributional statistics better matches the purchase prediction task.

\textit{2. Randomization improves robustness.}
Abacus-R matches or improves over Abacus on the private dataset and is comparable on the Taobao, suggesting that enforcing order-invariance
regularizes the objective
improving generalization.


\textit{3. Proposed objectives are complementary and stable.}
Mixing counting, masking, and contrastive tasks with MTL consistently exceeds the best single-task pretraining results 
being stable across the seeds and datasets/encoders reducing the standard deviation. This suggests that combining complementary inductive biases acts as an effective regularizer that enriches the learning.

\textit{4. Analysis of the Abacus Loss behavior.}
During pretraining, while decreasing, value of $\mathcal{L^{\text{Abacus}}}$ is quite high.
This is expected: transformers and GRU struggle with explicit counting\citep{yehudai2024can, weiss-etal-2018-practical}, being similar to distribution estimation.
Nevertheless, approximate moment-matching suffices to uplift downstream performance, as representations capture global distributional regularities without exact counts.



\textit{5. Objective misalignment and negative transfer.}
Objectives that enforce exact sequence fidelity either via reconstruction or the ordered future prediction (NEP, NKEHP, MSM) optimize a harder goal that involves prediction. NEP and MSM are moreover weakly aligned with downstream purchase prediction and require precise token order and timestamp recovery. This amplifies the exposure bias and error compounding, and are brittle to time noise, which translates into higher variance and frequent negative transfer. In contrast, in sequence distributional (counting) pretraining targets global statistics that are better matched to ranking and yields more reliable and stable gains across models and datasets.
 
\textit{6. Why GRU is better than BERT ?}
Observed sequences have a \emph{tiny} vocabulary (order $\sim\!10$ vs $\sim\!30{,}000$ in NLP) and are relatively short (order $\sim\!100$ tokens vs $\sim\!512$ in NLP).
In this regime, the Transformer advantages (global self-attention and long-range context) are underused, while its higher capacity can overfit and require more data/budget to stabilize. GRUs offer a stronger inductive bias for ordered, local temporal dynamics and count-like aggregation with far fewer parameters, leading to more stable training and better downstream AUC in our setting.

\vspace{-1em} 
\section{Conclusion}


We introduced Abacus, self-supervised objective of predicting empirical user event distribution. Across two datasets of user behavior logs and two encoders (GRU, BERT), Abacus consistently improves performance on downstream task. A hybrid, MTL approach, combining Abacus with MSM and BT tasks, further strengthens performance, indicating pretraining tasks complementarity. 

Future work will include scaling of count-aligned pretraining to longer event sequences and larger vocabularies, pretraining objectives focused on efficient event time encoding 
and application of the more flexible MTL architectures (e.g. MoSE \cite{qin2020multitask}).

\section{Ethical Considerations}

\paragraph{Data provenance and privacy.}
This work uses one private, large-scale behavioral dataset and one public dataset (Taobao).
The private dataset was accessed under a data-use agreement for research purposes.
It contains de-identified interaction logs consisting only of event types and timestamps; no direct identifiers (e.g., names, emails, device IDs) or content-level attributes were available to the authors.
We applied data minimization (event type, timestamp only), removed any residual quasi-identifiers where present, and restricted access to authorized researchers.
All analyses were conducted in aggregate, and we report only summary statistics and model performance.

\paragraph{Personal data and sensitive attributes.}
We did not use protected attributes (e.g., race, ethnicity, gender, religion) nor attempt to infer them.
Labels and features reflect user interactions, not personal characteristics.
We caution, however, that behavioral logs can still enable indirect profiling.
Any operational deployment should be preceded by a data protection impact assessment (DPIA) and appropriate legal review under applicable data protection laws and platform terms of service.

\paragraph{Potential harms and intended use.}
Our method can improve purchase prediction and, by extension, bidding efficiency in online advertising.
Potential risks include over-personalization, unwanted targeting, or reinforcement of filter bubbles.
We recommend guardrails in deployment: frequency capping, budget and exposure constraints, human oversight for campaign goals, transparent user controls (opt-outs where applicable), and continuous monitoring for unintended outcomes.

\paragraph{Fairness and bias.}
Because protected attributes were unavailable, we could not compute standard group fairness metrics.
To mitigate risk, we recommend pre-deployment audits using available proxies (e.g., geography or time-of-day segments), calibration monitoring across segments, and periodic counterfactual or perturbation testing to detect disparate error rates.
Future work should evaluate fairness with appropriate consented labels or privacy-preserving auditing protocols.

\paragraph{Security and confidentiality.}
Model artifacts trained on private data were not released.
All experiments on the public dataset use only public data and anonymized code.
No attempts were made to re-identify users or link records across sources.

\paragraph{Reproducibility and transparency.}
To support verification without exposing private data, we will release code, configuration files, and scripts to reproduce all public-dataset experiments, including data preprocessing and hyperparameters.
We also report distributional diagnostics (entropy, effective vocabulary size, Gini--Simpson) to clarify when the method is likely to be beneficial.

\paragraph{Environmental considerations.}
Training used two NVIDIA V100 GPUs and modest epochs; we prioritized early stopping and reuse of pretrained checkpoints to limit compute.
We encourage reporting energy use and preferring smaller backbones when performance is comparable.

\paragraph{Scope and limitations.}
Abacus aligns representations with distributional (counting) signals and is most effective when event-type distributions are rich.
The method does not guarantee exact counting and should not be used to infer sensitive attributes or make high-stakes decisions without human oversight and additional safeguards.

\bibliographystyle{ACM-Reference-Format}
\balance
\bibliography{biblio}

@inproceedings{zolna2017user,
  title     = {User2Vec: User Modeling Using LSTM Networks},
  author    = {{\.Z}o{\l}na, Konrad and Roma{\'n}ski, Bart{\l}omiej},
  booktitle = {Proceedings of the AAAI Conference on Artificial Intelligence},
  volume    = {31},
  number    = {1},
  year      = {2017}
}

@article{devlin2018bert,
  title   = {BERT: Pre-training of Deep Bidirectional Transformers for Language Understanding},
  author  = {Devlin, Jacob and Chang, Ming-Wei and Lee, Kenton and Toutanova, Kristina},
  journal = {arXiv preprint arXiv:1810.04805},
  year    = {2018}
}

@article{zbontar2021barlow,
  title   = {Barlow Twins: Self-Supervised Learning via Redundancy Reduction},
  author  = {Zbontar, Jure and Jing, Li and Misra, Ishan and LeCun, Yann and Deny, St{\'e}phane},
  journal = {arXiv preprint arXiv:2103.03230},
  year    = {2021}
}

@inproceedings{singh2023adload,
  author       = {Abhishek Singh and Rohit Gupta and Gaurav Pandey and others},
  title        = {Ad-Load Balancing via Off-policy Learning in a Content Marketplace},
  booktitle    = {Proceedings of the ACM Web Conference (WWW)},
  year         = {2023},
  url          = {https://arxiv.org/abs/2302.07888}
}

@inproceedings{zhou2018atrank,
  author       = {Fan Zhou and Runze Wu and Zhiyuan Xu and others},
  title        = {ATRank: An Attention-Based User Behavior Modeling Framework for Recommendation},
  booktitle    = {Proceedings of the AAAI Conference on Artificial Intelligence},
  year         = {2018},
  url          = {https://arxiv.org/abs/1711.06632}
}

@misc{pinterest2018aperture,
  author       = {Pinterest Engineering},
  title        = {Building a Real-time User Action Counting System for Ads},
  year         = {2018},
  howpublished = {\url{https://medium.com/pinterest-engineering/building-a-real-time-user-action-counting-system-for-ads-8ac54e569fe}},
}

@misc{snap2022robusta,
  author       = {Snap Inc. Engineering},
  title        = {Speed Up Feature Engineering for Recommendation Systems},
  year         = {2022},
  howpublished = {\url{https://eng.snap.com/speed-up-feature-engineering-for-recommendation-systems}},
}

@inproceedings{fu2023robust,
  title={Robust user behavioral sequence representation via multi-scale stochastic distribution prediction},
  author={Fu, Chilin and Wu, Weichang and Zhang, Xiaolu and Hu, Jun and Wang, Jing and Zhou, Jun},
  booktitle={Proceedings of the 32nd ACM International Conference on Information and Knowledge Management},
  pages={4567--4573},
  year={2023}
}

@inproceedings{lang2017understanding,
  title={Understanding consumer behavior with recurrent neural networks},
  author={Lang, Tobias and Rettenmeier, Matthias},
  booktitle={Workshop on Machine Learning Methods for Recommender Systems},
  year={2017}
}

@article{wu2020ptum,
  title={PTUM: Pre-training user model from unlabeled user behaviors via self-supervision},
  author={Wu, Chuhan and Wu, Fangzhao and Qi, Tao and Lian, Jianxun and Huang, Yongfeng and Xie, Xing},
  journal={arXiv preprint arXiv:2010.01494},
  year={2020}
}

@inproceedings{wu2022userbert,
  title={Userbert: Pre-training user model with contrastive self-supervision},
  author={Wu, Chuhan and Wu, Fangzhao and Qi, Tao and Huang, Yongfeng},
  booktitle={Proceedings of the 45th International ACM SIGIR Conference on Research and Development in Information Retrieval},
  pages={2087--2092},
  year={2022}
}

@misc{liao2020effectiveness,
  title={On the Effectiveness of Self-supervised Pre-training for Modeling User Behavior Sequences},
  author={Liao, Yiping},
  year={2020},
  publisher={AdKDD}
}

@article{liu2025enhancing,
  title={Enhancing User Sequence Modeling through Barlow Twins-based Self-Supervised Learning},
  author={Liu, Yuhan and Ning, Lin and Wu, Neo and Singhal, Karan and Mansfield, Philip Andrew and Berlowitz, Devora and Prakash, Sushant and Green, Bradley},
  journal={arXiv preprint arXiv:2505.00953},
  year={2025}
}

@inproceedings{fang2024general,
  title={General-Purpose User Modeling with Behavioral Logs: A Snapchat Case Study},
  author={Fang, Qixiang and Zhou, Zhihan and Barbieri, Francesco and Liu, Yozen and Neves, Leonardo and Nguyen, Dong and Oberski, Daniel and Bos, Maarten and Dotsch, Ron},
  booktitle={Proceedings of the 47th International ACM SIGIR Conference on Research and Development in Information Retrieval},
  pages={2431--2436},
  year={2024}
}

@inproceedings{sun2019bert4rec,
  title={BERT4Rec: Sequential recommendation with bidirectional encoder representations from transformer},
  author={Sun, Fei and Liu, Jun and Wu, Jian and Pei, Changhua and Lin, Xiao and Ou, Wenwu and Jiang, Peng},
  booktitle={Proceedings of the 28th ACM international conference on information and knowledge management},
  pages={1441--1450},
  year={2019}
}

@inproceedings{abbattista2024enhancing,
  title={Enhancing sequential music recommendation with personalized popularity awareness},
  author={Abbattista, Davide and Anelli, Vito Walter and Di Noia, Tommaso and Macdonald, Craig and Petrov, Aleksandr Vladimirovich},
  booktitle={Proceedings of the 18th ACM Conference on Recommender Systems},
  pages={1168--1173},
  year={2024}
}

@inproceedings{zhao2022resetbert4rec,
  title={RESETBERT4Rec: A pre-training model integrating time and user historical behavior for sequential recommendation},
  author={Zhao, Qihang},
  booktitle={Proceedings of the 45th international ACM SIGIR conference on research and development in information retrieval},
  pages={1812--1816},
  year={2022}
}

@article{gligorijevic2019time,
  title={Time-aware prospective modeling of users for online display advertising},
  author={Gligorijevic, Djordje and Gligorijevic, Jelena and Flores, Aaron},
  journal={arXiv preprint arXiv:1911.05100},
  year={2019}
}

@article{brown2020language,
  title={Language models are few-shot learners},
  author={Brown, Tom and Mann, Benjamin and Ryder, Nick and Subbiah, Melanie and Kaplan, Jared D and Dhariwal, Prafulla and Neelakantan, Arvind and Shyam, Pranav and Sastry, Girish and Askell, Amanda and others},
  journal={Advances in neural information processing systems},
  volume={33},
  pages={1877--1901},
  year={2020}
}

@inproceedings{chen2020simple,
  title={A simple framework for contrastive learning of visual representations},
  author={Chen, Ting and Kornblith, Simon and Norouzi, Mohammad and Hinton, Geoffrey},
  booktitle={International conference on machine learning},
  pages={1597--1607},
  year={2020},
  organization={PmLR}
}

@inproceedings{devlin2019bert,
  title={Bert: Pre-training of deep bidirectional transformers for language understanding},
  author={Devlin, Jacob and Chang, Ming-Wei and Lee, Kenton and Toutanova, Kristina},
  booktitle={Proceedings of the 2019 conference of the North American chapter of the association for computational linguistics: human language technologies, volume 1 (long and short papers)},
  pages={4171--4186},
  year={2019}
}

@inproceedings{kang2018self,
  title={Self-attentive sequential recommendation},
  author={Kang, Wang-Cheng and McAuley, Julian},
  booktitle={2018 IEEE international conference on data mining (ICDM)},
  pages={197--206},
  year={2018},
  organization={IEEE}
}

@inproceedings{zhou2020s3,
  title={S3-rec: Self-supervised learning for sequential recommendation with mutual information maximization},
  author={Zhou, Kun and Wang, Hui and Zhao, Wayne Xin and Zhu, Yutao and Wang, Sirui and Zhang, Fuzheng and Wang, Zhongyuan and Wen, Ji-Rong},
  booktitle={Proceedings of the 29th ACM international conference on information \& knowledge management},
  pages={1893--1902},
  year={2020}
}

@inproceedings{han2024enhancing,
  title={Enhancing CTR Prediction through Sequential Recommendation Pre-training: Introducing the SRP4CTR Framework},
  author={Han, Ruidong and Li, Qianzhong and Jiang, He and Li, Rui and Zhao, Yurou and Li, Xiang and Lin, Wei},
  booktitle={Proceedings of the 33rd ACM International Conference on Information and Knowledge Management},
  pages={3777--3781},
  year={2024}
}

@inproceedings{du2023frequency,
  title={Frequency enhanced hybrid attention network for sequential recommendation},
  author={Du, Xinyu and Yuan, Huanhuan and Zhao, Pengpeng and Qu, Jianfeng and Zhuang, Fuzhen and Liu, Guanfeng and Liu, Yanchi and Sheng, Victor S},
  booktitle={Proceedings of the 46th international ACM SIGIR conference on research and development in information retrieval},
  pages={78--88},
  year={2023}
}

@article{yehudai2024can,
  title={When Can Transformers Count to n?},
  author={Yehudai, Gilad and Kaplan, Haim and Ghandeharioun, Asma and Geva, Mor and Globerson, Amir},
  journal={arXiv preprint arXiv:2407.15160},
  year={2024}
}

@inproceedings{weiss-etal-2018-practical,
  title     = {On the Practical Computational Power of Finite Precision RNNs for Language Recognition},
  author    = {Weiss, Gail and Goldberg, Yoav and Yahav, Eran},
  booktitle = {Proceedings of the 56th Annual Meeting of the Association for Computational Linguistics (Volume 2: Short Papers)},
  year      = {2018},
  month     = jul,
  address   = {Melbourne, Australia},
  publisher = {Association for Computational Linguistics},
  url       = {https://aclanthology.org/P18-2117},
  doi       = {10.18653/v1/P18-2117}
}

@inproceedings{pancha2022pinnerformer,
  title={Pinnerformer: Sequence modeling for user representation at pinterest},
  author={Pancha, Nikil and Zhai, Andrew and Leskovec, Jure and Rosenberg, Charles},
  booktitle={Proceedings of the 28th ACM SIGKDD conference on knowledge discovery and data mining},
  pages={3702--3712},
  year={2022}
}

@misc{wang2023bert4ctrefficientframeworkcombine,
      title={BERT4CTR: An Efficient Framework to Combine Pre-trained Language Model with Non-textual Features for CTR Prediction}, 
      author={Dong Wang and Kavé Salamatian and Yunqing Xia and Weiwei Deng and Qi Zhiang},
      year={2023},
      eprint={2308.11527},
      archivePrefix={arXiv},
      primaryClass={cs.CL},
      url={https://arxiv.org/abs/2308.11527}, 
}

@inproceedings{qin2020multitask,
  title={Multitask mixture of sequential experts for user activity streams},
  author={Qin, Zhen and Cheng, Yicheng and Zhao, Zhe and Chen, Zhe and Metzler, Donald and Qin, Jingzheng},
  booktitle={Proceedings of the 26th ACM SIGKDD international conference on knowledge discovery \& data mining},
  pages={3083--3091},
  year={2020}
}

\end{document}